\documentclass{article}

\usepackage{PRIMEarxiv}

\usepackage[utf8]{inputenc}
\usepackage[T1]{fontenc}
\usepackage{hyperref}
\usepackage{url}
\usepackage{booktabs}
\usepackage{amsfonts}
\usepackage{nicefrac}
\usepackage{microtype}
\usepackage{graphicx}
\usepackage{fancyhdr}
\usepackage{tabularx}
\usepackage{longtable}
\usepackage{arabtex}
\usepackage{utf8}
\setcode{utf8}

\graphicspath{{media/}}

\title{Algerian Dialect \\
\Large A Sentiment-Annotated YouTube Comment Dataset in Algerian Arabic Dialect
\thanks{\textit{Citation:} Benmounah, Z., Boulesnane, A., Fadheli, A., \& Khial, M. (2023). Sentiment Analysis on Algerian Dialect with Transformers. \textit{Applied Sciences}, 13(20), 11157. \href{https://doi.org/10.3390/app132011157}{https://doi.org/10.3390/app132011157}}}

\author{
  Zakaria Benmounah \\
  LISIA Laboratory\\
  Abdelhamid Mehri University Constantine 02\\
  Constantine, Algeria \\
  \texttt{zakaria.benmounah@univ-constantine2.dz} \\
  \And
  Abdennour Boulesnane \\
  Faculty of Medicine, BIOSTIM Lab \\
  Constantine 3 University \\
  Constantine, Algeria \\
  \texttt{aboulesnane@univ-constantine3.dz}\\
   \And
 Abdeladim Fadheli, and Mustapha Khial \\
NTIC Faculty\\
  Abdelhamid Mehri University Constantine 02\\
  Constantine, Algeria
}

\renewenvironment{abstract}
{
  \small
  \begin{center}
  \bfseries Abstract
  \end{center}
  \begin{quote}
}
{
  \end{quote}
}

\begin{document}
\maketitle

\begin{abstract}
We present \textbf{Algerian Dialect}, a large-scale sentiment-annotated dataset consisting of 45,000 YouTube comments written in Algerian Arabic dialect. The comments were collected from more than 30 Algerian press and media channels using the YouTube Data API. Each comment is manually annotated into one of five sentiment categories: very negative, negative, neutral, positive, and very positive. In addition to sentiment labels, the dataset includes rich metadata such as collection timestamps, like counts, video URLs, and annotation dates. This dataset addresses the scarcity of publicly available resources for Algerian dialect and aims to support research in sentiment analysis, dialectal Arabic NLP, and social media analytics. The dataset is publicly available on Mendeley Data under a CC BY 4.0 license at \href{https://doi.org/10.17632/zzwg3nnhsz.2}{10.17632/zzwg3nnhsz.2}.
\end{abstract}

\keywords{Algerian Dialect, Sentiment Analysis, YouTube Comments, Arabic NLP, Annotated Dataset, Low-Resource Languages}

\section{Introduction}
The rapid growth of social media platforms has generated vast amounts of user-generated textual data that reflect public opinion, emotions, and reactions to real-world events \cite{Sun2024}. Sentiment analysis has become a central task in natural language processing (NLP) with applications in political analysis, marketing, social sciences, and media monitoring \cite{Wankhade2022}. While significant progress has been achieved for high-resource languages, many Arabic dialects remain severely under-resourced.

Algerian Arabic is one of the most widely spoken Arabic dialects, yet it poses unique challenges for NLP due to its strong divergence from Modern Standard Arabic (MSA), extensive lexical borrowing from French and Berber languages, and frequent code-switching \cite{Harrat2016}. Existing Arabic sentiment datasets often focus on MSA or use coarse-grained sentiment labels, limiting their applicability to real-world dialectal content.

To address these limitations, we introduce a large-scale, manually annotated dataset of Algerian dialect YouTube comments. By focusing on press and media channels, the dataset captures authentic public discourse on political, social, and economic topics, making it particularly valuable for sentiment analysis and opinion mining tasks.

\section{Value of the Data}
\begin{itemize}
  \item Provides one of the largest publicly available sentiment-annotated corpora in Algerian Arabic dialect.
  \item Employs a fine-grained five-class sentiment scheme, enabling nuanced sentiment modeling.
  \item Captures real-world informal language, including slang, emojis, and code-switching.
  \item Includes rich metadata supporting temporal and engagement-based analyses.
  \item Facilitates benchmarking and comparison of sentiment models for dialectal Arabic.
\end{itemize}

\section{Data Collection}
Data collection was performed using the official YouTube Data API. Comments were extracted from over 30 Algerian press and media channels selected for their relevance and audience engagement. The collection process prioritized comments written in Arabic script and commonly used Algerian dialect expressions.

To respect user privacy, no personal identifiers were stored. Only publicly available comment content and non-sensitive metadata were retained. Spam, duplicated content, and non-relevant entries were filtered during preprocessing.

\section{Annotation Protocol}
The sentiment annotation process was conducted manually by native Algerian Arabic speakers with linguistic expertise. The objective was to capture fine-grained emotional polarity expressed in informal online discourse. To this end, a five-level sentiment scale was adopted instead of a coarse binary or ternary scheme.

Each comment was assigned to exactly one sentiment category, ranging from very negative to very positive. The definition of each label is summarized in Table~\ref{tab:sentiment_labels}. This labeling scheme allows the dataset to support nuanced sentiment modeling, including intensity-aware classification and ordinal sentiment analysis.
\begin{table}[h]
\centering
\caption{Sentiment Label Definitions}
\label{tab:sentiment_labels}
\begin{tabular}{cl}
\toprule
\textbf{Label} & \textbf{Sentiment Meaning} \\
\midrule
0 & Very Negative \\
1 & Negative \\
2 & Neutral \\
3 & Positive \\
4 & Very Positive \\
\bottomrule
\end{tabular}
\end{table}

During the annotation process, annotators relied on contextual cues, lexical choices, emoji usage, and pragmatic meaning to determine sentiment polarity. Ambiguous or context-dependent comments were reviewed collectively, and disagreements were resolved through discussion. This protocol helped ensure annotation consistency and reduced subjectivity, resulting in a high-quality gold-standard dataset suitable for supervised learning and benchmarking tasks.

\section{Dataset Description}
Each record in the dataset is composed of the original comment text along with sentiment annotations and auxiliary metadata. The inclusion of metadata enables not only sentiment classification but also temporal, engagement-based, and content-source analyses. Table~\ref{tab:dataset_fields} provides a detailed description of all fields included in the released dataset.
\newpage
\begin{table}[h]
\centering
\caption{Description of the Algerian Dialect Dataset Fields}
\label{tab:dataset_fields}
\begin{tabularx}{\linewidth}{lc}
\toprule
\textbf{Field Name} & \textbf{Description} \\
\midrule
text & Raw YouTube comment text written in Algerian Arabic dialect \\
label & Sentiment annotation label ranging from 0 (very negative) to 4 (very positive) \\
last\_updated & Date and time when the comment was collected via the YouTube API \\
like\_count & Number of likes received by the comment at collection time \\
video\_url & URL of the YouTube video from which the comment was extracted \\
annotation\_datetime & Timestamp indicating when the sentiment annotation was performed \\
\bottomrule
\end{tabularx}
\end{table}
The dataset is distributed in a structured tabular format to facilitate direct use in machine learning pipelines. The raw comment text is preserved without aggressive normalization in order to retain linguistic phenomena such as spelling variation, code-switching, and informal expressions, which are characteristic of Algerian dialect in online platforms.

\begin{figure}[h!]
\centering
\includegraphics[width=\linewidth]{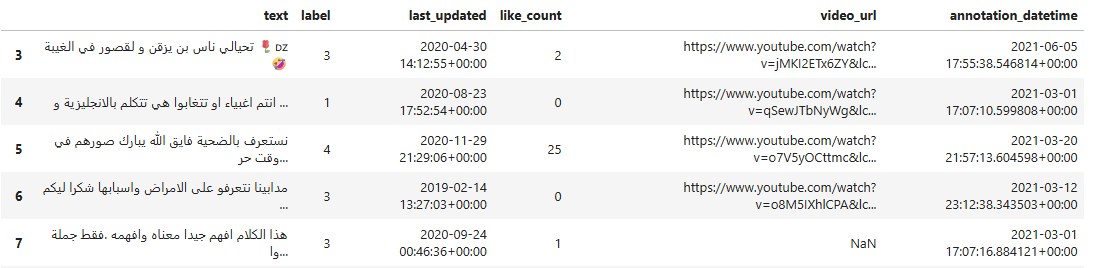}
\caption{Preview of selected entries from the Algerian Dialect dataset showing comment text and associated sentiment labels.}
\label{fig:dataset_preview}
\end{figure}

To provide a concrete illustration of the dataset content and structure, Figure~\ref{fig:dataset_preview} shows a small sample of comments extracted from the dataset along with their corresponding sentiment labels. The displayed examples highlight typical characteristics of Algerian dialect used on social media, including informal expressions, dialectal vocabulary, and varied sentiment polarity.

As shown in Table~\ref{tab:sentiment_distribution}, the dataset exhibits a noticeable imbalance among sentiment categories, with negative and positive sentiments being more frequent than very negative and very positive ones. This distribution mirrors real-world user engagement on social media platforms, where extreme sentiments are typically less frequent. While this imbalance may require the use of appropriate evaluation metrics or class-weighting strategies, it also enhances the realism of the dataset and makes it well suited for studying sentiment dynamics in authentic online environments.
\begin{table}[h!]
\centering
\caption{Distribution of Sentiment Classes in the Dataset}
\label{tab:sentiment_distribution}
\begin{tabular}{lcc}
\toprule
\textbf{Sentiment Class} & \textbf{Number of Comments} & \textbf{Percentage (\%)} \\
\midrule
Very Negative & 1,543 & 3.43 \\
Negative & 14,414 & 32.03 \\
Neutral & 10,093 & 22.43 \\
Positive & 16,747 & 37.21 \\
Very Positive & 2,203 & 4.90 \\
\midrule
\textbf{Total} & \textbf{45,000} & \textbf{100.00} \\
\bottomrule
\end{tabular}
\end{table}

The dataset presented here was originally used in \cite{benmounah2023} to evaluate transformer-based models for sentiment analysis, demonstrating its suitability for deep learning approaches.
\section{Linguistic Characteristics}
The dataset reflects authentic Algerian online communication. Approximately 15\% of comments contain emojis, which often convey sentiment polarity. Many comments exhibit code-switching between Arabic and Latin scripts, as well as lexical borrowing from French. These characteristics increase the dataset’s realism while posing challenges for NLP models.

\section{Preprocessing and Baseline Usage}
In the associated study \cite{benmounah2023}, several preprocessing strategies were evaluated, including Arabic normalization, emoji tokenization, and URL replacement with placeholder tokens. The dataset was successfully used to train transformer-based models, achieving competitive accuracy and F1-scores, confirming its suitability for advanced NLP applications.

\section{Applications and Use Cases}
Potential applications include:
\begin{itemize}
  \item Sentiment classification in Algerian dialect.
  \item Fine-tuning large language models for dialectal Arabic.
  \item Public opinion and media analysis.
  \item Cross-dialect and cross-lingual transfer learning.
  \item Sociolinguistic studies of Algerian online discourse.
\end{itemize}

\section{Data Availability and Citation}
The dataset presented in this study is publicly available on Mendeley Data at
\href{https://data.mendeley.com/datasets/zzwg3nnhsz/2}{https://data.mendeley.com/datasets/zzwg3nnhsz/2}.
The dataset is released under the Creative Commons Attribution 4.0 (CC BY 4.0) license,
allowing unrestricted use, distribution, and reproduction, provided that the original
authors are properly cited.

If you use this dataset in your research, please cite the following original work:

\begin{quote}
Benmounah, Z., Boulesnane, A., Fadheli, A., \& Khial, M. (2023).
\textit{Sentiment Analysis on Algerian Dialect with Transformers}.
Applied Sciences, 13(20), 11157.
\href{https://doi.org/10.3390/app132011157}{https://doi.org/10.3390/app132011157}
\end{quote}

\section*{Acknowledgments}
We thank the annotators and the NLP research community for their valuable feedback and support.

\bibliographystyle{unsrt}

\end{document}